\documentclass[11pt]{article}

\usepackage[final]{acl}

\usepackage{times}
\usepackage{latexsym}

\usepackage[T1]{fontenc}
\usepackage{booktabs}
\usepackage{siunitx}
\usepackage{amsmath}

\usepackage[utf8]{inputenc}

\usepackage{microtype}
\usepackage{enumitem}

\usepackage{xcolor}
\usepackage[most]{tcolorbox}

\newcommand{\pnote}[1]{\textcolor{blue}{\,(#1)}}
\newtcolorbox{promptbox}[1]{breakable,colback=black!3,colframe=black!55,colbacktitle=black!15,coltitle=black,fonttitle=\bfseries\small,fontupper=\small,title={#1},boxrule=0.5pt,arc=1pt,left=5pt,right=5pt,top=4pt,bottom=4pt}

\usepackage{inconsolata}

\usepackage{graphicx}

\title{Chiaroscuro for Emotions: A Contrastive Emotion Benchmark Grounded in Appraisal Theory}

\author{Divyesh Bommana, Mohammad Saim, Tianyu Jiang \\
        University of Cincinnati \\
        \texttt{bommandh@mail.uc.edu, saimmd@mail.uc.edu, tianyu.jiang@uc.edu}}

\begin{document}
\maketitle
\begin{abstract}
Emotion recognition benchmarks often predict one emotion per text, missing many real-world scenarios where two people arrive at opposing emotions from a single shared event. For example, a child kicks the seat in front of her in excitement while the passenger ahead grows angry. We introduce \textsc{Chiaro}, a $1{,}000$ human-annotated sentence benchmark for contrastive emotion inference grounded in appraisal theory. Each scene describes one causal trigger eliciting a positive emotion in one person and a negative emotion in the other, drawn from a ten-class taxonomy. We benchmark seven frontier LLMs and four off-the-shelf emotion classifiers. The strongest LLM reaches $67.3$ macro-$F_1$, well below human agreement, while existing emotion classifiers score near chance. Beyond evaluation, \textsc{Chiaro} also serves as a training signal. When combined with an existing emotion corpus, the resulting downstream classifier improves on \textsc{Chiaro} itself and on six of ten external emotion benchmarks, which positions our dataset as a complementary signal for emotion recognition.
\end{abstract}

\section{Introduction}

When two people share a single event, they can often arrive at opposing emotions. Each person reacts to a different aspect of the same situation, and the text rarely names either feeling outright. For example, a surprise promotion announced in front of the whole team may fill one engineer with pride at the recognition, while the colleague who had been quietly competing for the same role feels their stomach drop as the news lands. Neither emotion is stated, yet both are inferable from the situation alone. Understanding both emotions is the unit of analysis that many tasks need: conversational systems that mediate interpersonal disputes \citep{yeo-jaidka-2025-beyond}, story generation that must render each character’s reaction to a scene, and multi-party dialogue analysis where emotions routinely diverge within a shared event \citep{poria-etal-2019-meld} or even account for how affect shapes ethical judgments of a situation~\citep{saim-jiang-2026-emotions}. 

\begin{figure}[t]
    \centering
    \includegraphics[width=0.95\columnwidth]{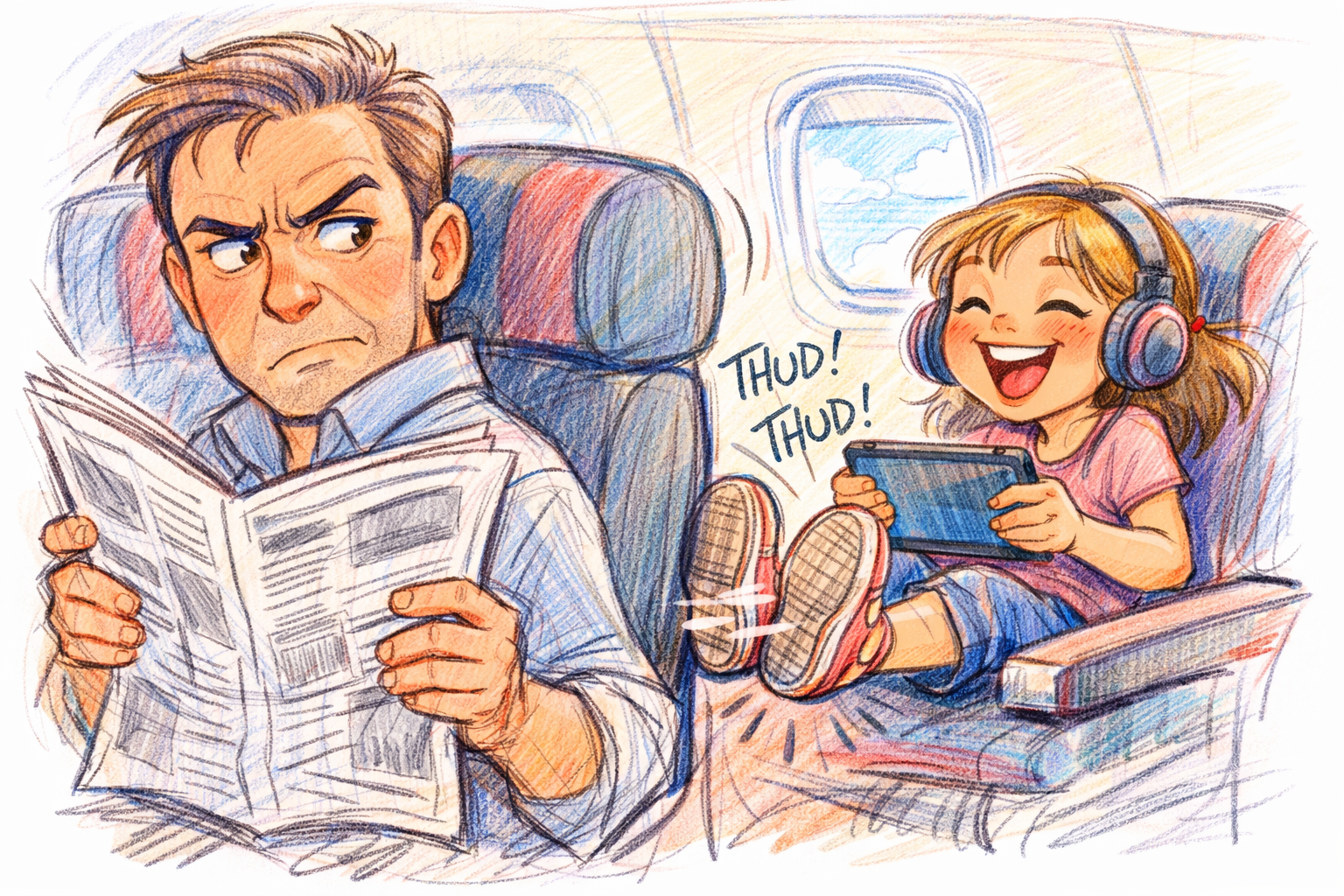}
    \caption{Contrastive emotions in a shared scene. A child seated behind gleefully kicks the seat while playing on a tablet, whereas the man in front turns back with visible annoyance.}
    \label{fig:intro}
\end{figure}

From GoEmotions \citep{demszky-etal-2020-goemotions} to its recent multilingual and culturally-grounded successors \citep{muhammad-etal-2025-brighter,belay-etal-2025-culemo}, fine-grained labeled corpora have grown substantially in scale and coverage. However, the prediction target remains the emotion of a single person in isolation, and a strong baseline can often be built from a single affective keyword \citep{sabour-etal-2024-emobench}. The Implicit Emotion Shared Task \citep{klinger-etal-2018-iest} partially addresses this limitation by removing the explicit affect word, but is limited to a single person experiencing the emotion. We propose a dataset that targets the joint, opposed-valence reading illustrated in Figure~\ref{fig:intro}, where two people in a shared scene have contrasting emotions tied to a shared cause. The proposed 1,000-sentence benchmark contains emotions from both valences (positive and negative) drawn from a balanced ten-class taxonomy and grounded in appraisal theory. Further, our benchmark shows a substantial gap between the best-performing frontier LLM and human agreement. 

We highlight the evidence of our grounding. Appraisal theory states that emotions are not produced by events directly but by an agent’s evaluation of events along dimensions such as goal congruence, agency, and certainty \citep{smith-ellsworth-1985,roseman-1996,ortony1988cognitive,ellsworth-scherer-2003-appraisal,moors-etal-2013-appraisal}. Two individuals witnessing the same event under different goals or different agency can arrive at opposed emotions. For example, an unannounced snow day delights the kids and dismays the working parents scrambling for last-minute childcare. This framing is important for structuring a contrastive sentence. We introduce \textbf{\textsc{Chiaro}},\footnote{From \emph{chiaroscuro}, the painterly technique of rendering strong light--dark contrast on a single canvas (Caravaggio, Rembrandt, Vermeer); a metaphor for opposing emotions arising from one shared event.} a benchmark dataset of 1{,}000 sentences where each sentence describes a single causal trigger eliciting a positive emotion in one agent and a negative emotion in the other, drawn from a balanced ten-class taxonomy. We avoid explicit use of affect words, and emotion must be inferred from situational context alone, with human annotations for both agents in each scene. Overall, our contributions are threefold:
\begin{enumerate}[itemsep=2pt, topsep=4pt]
  \item We introduce \textsc{Chiaro},\footnote{\url{https://github.com/cincynlp/Chiaro}} a 1{,}000-sentence benchmark for two-person contrastive emotion inference in a single shared event, grounded in appraisal theory.
  \item We benchmark seven frontier LLMs and four off-the-shelf emotion classifiers on \textsc{Chiaro}, showing that even the strongest frontier model falls well below human agreement and that existing single-agent emotion classifiers transfer to the task only at chance level.
  \item We establish \textsc{Chiaro} as a complementary training resource for existing emotion classifiers. A RoBERTa-large fine-tuned on the union of \textsc{Chiaro} and a matched-size slice of an existing emotion dataset like GoEmotions beats either source alone on \textsc{Chiaro} and on six of ten external emotion benchmarks.
\end{enumerate}

\section{Related Works}
The study of emotions in NLP developed from early affective text classification and sentiment polarity benchmarks ~\citep{strapparava2007semeval, pang2008opinion, mohammad-etal-2018-semeval} into a broad research program spanning lexical resources ~\citep{mohammad2013crowdsourcing}, dimensional annotation frameworks~\citep{buechel2017emobank}, large-scale multi-label corpora ~\citep{demszky-etal-2020-goemotions, muhammad-etal-2025-brighter} and embodied inference ~\citep{zhuang-etal-2024-heart, duong-etal-2025-cheer, saim-etal-2025-anatomy}. Compositional approaches show that sentiment is not monolithic within a passage and that conflicting polarities can attach to distinct targets \citep{socher2013recursive, pontiki2014semeval}, while implicit emotion tasks have established that surface affect words are neither necessary nor sufficient for inference ~\citep{klinger-etal-2018-iest}. Emotion recognition in conversations extended this framework to multi-speaker settings, where models must track the affective state across turns and infer emotion from social context ~\citep{li-etal-2017-dailydialog, poria-etal-2019-meld, rashkin-etal-2019-towards, ghosal2020cosmic}. Causal reasoning over emotion extends this to emotion-cause pair extraction ~\citep{xia2019emotion, poria2020recognizing} and shared tasks on conversation-level cause analysis ~\citep{wang2024semeval}, which require models to jointly identify an emotion and the event that triggered it, motivating the cause-span objective we adopt. A related line of work attaches affective polarity to events rather than to speakers~\citep{ding2018event, zhuang-etal-2020-affective}. Recent benchmarks probe whether LLMs genuinely reason about emotion or merely match surface patterns ~\citep{sabour-etal-2024-emobench, zhao2024both}. However, most evaluations in the emotion recognition space retain only the single-agent framing. The core aspect examined is which label applies to a single speaker, not how opposing valence is distributed between two agents who share a causal trigger.

Appraisal theories~\citep{scherer2001appraisal, ortony1988cognitive} account for the mechanism by which the same external event elicits different emotions in different agents, because each agent evaluates that event against their own goals and concerns. This formalizes the contrastive emotion setup as to why a single action yields delight in one agent and irritation in another. The research on co-occurring and mixed emotions confirms that opposite-valence states are not mutually exclusive and resist reliable recovery from surface form ~\citep{berrios2015eliciting, larsen2001people}. This validates the non-trivial inference challenge our dataset poses.

Work on interpersonal emotion regulation ~\citep{hatfield1993emotional} formalizes the directional influence from one agent’s expressive behavior to another’s affective response. We employ a similar framework while designing the shared space between our agents in each scenario.
Research in contrastive affect is sparse and has primarily appeared in aspect-based sentiment analysis ~\citep{pontiki2014semeval, schouten2015survey}, where conflicting
polarities attach to distinct opinion targets within a single document’s meaning. No existing dataset jointly requires a model to detect that two agents hold opposing valence and attribute the correct polarity to each agent by role. \textsc{CHIARO} targets this conjunction directly, providing paired physical and non-physical scenario variants for causal grounding and role-aware polarity attribution in the settings where current models most consistently fail.

\section{Task and Dataset Creation}
\label{sec:dataset}

\paragraph{Task definition.}
We formalize contrastive emotion inference as follows. Given a sentence describing a shared event involving two people (\emph{agents} A and B) together with each person's role, predict one emotion per person from a ten-class taxonomy comprising five positive and five negative classes. Each scene is constructed so that exactly one person’s emotion is positive and the other’s is negative. The prediction for a scene is therefore a paired assignment over the two people. The input contains no explicit affect words, so emotion must be inferred from situational context alone.

\paragraph{Motivation.}
Contrastive emotion inference is a novel evaluation target for emotion modeling. It is the joint prediction of two opposed emotions held by co-agents whose reactions diverge from a single shared trigger. Single-agent corpora such as GoEmotions \citep{demszky-etal-2020-goemotions}, ISEAR \citep{scherer-wallbott-1994-isear}, and EmpatheticDialogues \citep{rashkin-etal-2019-towards} treat each text segment as one emotion held by one writer or speaker; multi-party dialogue corpora such as MELD \citep{poria-etal-2019-meld} and DailyDialog \citep{li-etal-2017-dailydialog} label one emotion per utterance per speaker. Neither captures the joint, opposed-valence reading our task demands. Appraisal theory directly predicts this case: two agents witnessing the same event with different goals or agency can arrive at opposite emotions. \textsc{Chiaro} presents the prediction as both a benchmark dataset and an evaluation task.

\textsc{Chiaro} is constructed by converting short subreddit narratives into controlled, two-agent scenes that (i) exhibit opposing emotional valence across agents and (ii) require emotion inference from situational context rather than explicit affect vocabulary. The construction pipeline has three stages: source-narrative selection, two-stage scene generation, and automated validation and correction.

\subsection{Emotion Taxonomy}
\label{sec:taxonomy}

We label each agent with one of ten emotions, partitioned into five positive (joy, pride, relief, gratitude, excitement) and five negative (anger, sadness, fear, disgust, embarrassment) classes.

The taxonomy is derived from the GoEmotions \citep{demszky-etal-2020-goemotions} dataset. We select the five \emph{most distinct} emotions per positive--negative polarity by applying the following criteria to the dataset. First, hierarchical clustering analysis shows that several emotions form intensity pairs or near-synonym clusters (e.g., anger/annoyance, fear/nervousness, sadness/grief). We retain the more reliable representative for each cluster based on arousal. We also discard overlapping appraisal structures (e.g., \emph{love}, \emph{caring}, and \emph{admiration} as they all share \emph{gratitude}'s other-directed-positive cell). Second, each retained emotion occupies a distinct cell along the agency, certainty, and control dimensions of appraisal-theoretic models \citep{smith-ellsworth-1985,roseman-1996,ortony1988cognitive}. We also considered alternative taxonomies. A popular alternative is Ekman's six basic emotions, but it provides too few same-valence categories (only one positive emotion) for balanced contrastive scenes. Plutchik's eight emotions include valence-ambiguous categories (\emph{surprise}, \emph{anticipation}) that are incompatible with the opposed-valence design. Moreover, as mentioned above, the full GoEmotions inventory of 27 emotion categories (28 with \emph{neutral}) contains near-synonym clusters that make balanced two-agent generation and reliable annotation infeasible at our scale.

\begin{table*}[t]
\centering
\small
\setlength{\tabcolsep}{6pt}
\renewcommand{\arraystretch}{1.3}
\begin{tabular}{p{0.62\textwidth}ll}
\toprule
\textbf{Sentence} & \textbf{Agent A} & \textbf{Agent B} \\
\midrule
At a quiet corner table in the coffee shop, Maya's laptop shows an early acceptance email, and she reaches for her phone to call her mom, while Jordan, who thought they were applying as a pair, sees it and confronts her for submitting without him.
 & Jordan: \emph{anger} & Maya: \emph{excitement} \\
\midrule
When the door wedge catches with a loud thump and the door won't swing inward, the daughter inside the bedroom knows it will stay shut, while her mother in the hallway thinks someone is forcing the door and may get in before she can help.
 & the daughter: \emph{relief} & the mother: \emph{fear} \\
\midrule
At the store raffle board, the posted results list Maya as the winner and Lena as the runner-up, so Maya takes the prize voucher while Lena argues with the event staff about the listing.
 & Maya: \emph{joy} & Lena: \emph{anger} \\
\bottomrule
\end{tabular}
\caption{Three example \textsc{Chiaro} scenes spanning different emotion pairs. Each sentence describes a single shared event from which two agents arrive at opposed emotions; no explicit affect words appear in the sentence text.}
\label{tab:examples}
\end{table*}

Each retained emotion is then paired with a mandatory event or a trigger. The generated sentence must instantiate a situational feature so the emotion is recoverable from the source text (e.g., \emph{gratitude} requires an identifiable helper; \emph{relief} requires a prior threat that is then avoided). The full mapping and filtering from the GoEmotions taxonomy to \textsc{Chiaro} and the per-emotion mandatory triggers are listed in Appendix~\ref{app:triggers}.

\subsection{Source Narratives}
\label{sec:source}

We draw narrative inspiration from the \emph{r/AmItheAsshole} (AITA) subreddit, a long-running community where users post first-person accounts of interpersonal conflicts and seek moral judgment. We collect AITA posts via the Reddit API. Each post presents a self-contained interaction involving multiple participants mentioned in the story, a sequence of events, and implicit questions about the situation’s morality. We chose AITA over neutral story corpora (e.g., ROCStories) for two reasons. First, AITA posts are organized around interpersonal events with opposing affective stakes between participants, which is the structure required to generate contrastive emotion scenes. Second, AITA posts are dense in implicit appraisal cues (fairness, agency, harm, benefit) without naming the emotions themselves. Therefore, the genre itself models appraisal-based reasoning over situational evidence. Appendix~\ref{app:aita-keywords} describes the keyword-based post selection procedure.

\subsection{Two-Stage Generation}
\label{sec:pipeline}

Given a target emotion pair and an AITA Reddit post, we generate a \textsc{Chiaro} instance in two stages: \textsc{Draft} and \textsc{Render}. All generations are performed with OpenAI's \texttt{gpt-5.2}; decoding hyperparameters are reported in Appendix~\ref{app:prompts}.

\paragraph{Stage 1: Draft scene.}
The model produces a short draft scene with a one-sentence \texttt{setting}, two \texttt{agent\_roles}, and a 1--2-sentence \texttt{draft\_story} describing a concrete event that plausibly elicits opposing-valence emotions in the two agents. Each draft must satisfy five requirements:
(i) the two agents must hold opposed valence;
(ii) the scene must make clear \emph{why} each agent feels the way they do with coherent framing of the story;
(iii) the scene must avoid villain framings (e.g., theft, punishment, sabotage);
(iv) each agent’s emotion must instantiate the corresponding mandatory trigger; and (v) the draft must conform to one of six contrastive scenario types. Each type encodes a distinct causal structure linking the two agents’ outcomes. The full list of the scenario types is given in Appendix~\ref{app:scenarios}.

\paragraph{Stage 2: Paired render.}
From the draft scene, we generate two versions: a \emph{physical} and a \emph{non-physical}. The two versions share the agents and the underlying contrast but differ in the \emph{causal mode} of the triggering event:
\begin{itemize}
  \item \textbf{Physical:} the trigger involves contact, force, or object manipulation that directly changes one agent's situation (spilling, bumping, taking, breaking).
  \item \textbf{Non-physical:} the trigger is grounded in social or environmental cues without direct physical impact (overhearing, witnessing, knowing, announcing).
\end{itemize}

The two modes impose qualitatively different inferential demands. Physical triggers are concrete and immediately observable, whereas non-physical triggers require theory-of-mind reasoning about what each agent knows, perceives, or expects \citep{stein-1917-empathy,zahavi-2011-empathy}. Pairing both modes in the corpus lets us test whether a model’s emotion-attribution capacity transfers across causal modes or is biased toward one.

This motivation for a physical/non-physical categorization is grounded in the contrast between mechanical and experiential modes of causation. Stein's philosophy, as characterized by \citet{szanto-moran-2025-stein}, distinguishes \emph{mechanical} from \emph{experiential} causation. Mechanical causation is a tripartite chain of a \emph{verursachendes} (causing event), a \emph{verursachtes Geschehen} (caused event), and a mediating \emph{Ursache} (proper cause); experiential causation (\emph{Erlebniskausalit\"at}) is one in which two experiences jointly effectuate affective change without a separable mediating element. Physical scenes in \textsc{Chiaro} instantiate the former (the spilled coffee mechanically changes B’s situation), while non-physical scenes instantiate the latter (the overheard remark changes the other agent’s experience only via its meaning). Table~\ref{tab:examples} shows three example \textsc{Chiaro} scenes spanning different emotion pairs and causal modes.

\subsection{Validation of the Dataset}
\label{sec:validation}

A core design goal of \textsc{Chiaro} is to discourage trivial inference from explicit affect words. We enforce this through (a) lexical constraints applied to all natural-language fields and (b) a suite of programmatic validators applied to every generated instance. For example, instead of \emph{``Maya felt nervous as she waited for her exam result,''} a \textsc{Chiaro}-compliant rendering would describe the scene without naming the emotion: \emph{``Maya kept refreshing the portal every few seconds while the result loaded.''}

\paragraph{Lexical constraints.}
We exclude approximately seventy affect-bearing words and phrases from the natural-language sentence generation. The list covers explicit emotion adjectives (\emph{happy, sad, proud, angry, etc.}), and their morphological variants, stereotyped facial-expression descriptors (\emph{smiles, frowns, grins, glares, etc.}), and other high-leakage phrases (\emph{slumped shoulders, tight jaw, welling eyes, etc.}). The complete list is given in Appendix~\ref{app:banned}.

\paragraph{Validators and correction.}
Six programmatic checks are applied to each generated version, covering \emph{valence contrast}, \emph{lexical constraints}, \emph{text length}, \emph{person-reference count} (to avoid scene crowding), \emph{span consistency}, and \emph{role-head collision} (the two agents must be distinguishable from their role prefixes alone). Versions that fail one or more checks are returned to the model together with an explicit list of violations and a correction request; the repair loop is bounded by a small number of retries, and instances that repeatedly fail are discarded. Appendix~\ref{app:validators} gives full validator specifications.

\subsection{Human Annotation and Statistics}
\label{sec:annotation}

\begin{table}[t]
\centering
\small
\setlength{\tabcolsep}{10pt}
\begin{tabular}{lr@{\hspace{2em}}lr}
\toprule
\multicolumn{2}{c}{\textbf{Positive}} & \multicolumn{2}{c}{\textbf{Negative}} \\
\midrule
gratitude  & 22.5 & anger         & 25.3 \\
relief     & 22.1 & embarrassment & 21.0 \\
joy        & 20.1 & fear          & 19.9 \\
excitement & 18.7 & sadness       & 17.4 \\
pride      & 16.6 & disgust       & 16.4 \\
\bottomrule
\end{tabular}
\caption{Per-emotion ratio within polarity in \textsc{Chiaro}, measured over $1{,}000$ adjudicated gold labels per polarity (one per scene). Perfect balance is $20\%$ per class. Each scene contributes one positive and one negative slot.}
\label{tab:stats}
\end{table}

\begin{table*}[t]
\centering\small
\setlength{\tabcolsep}{8pt}
\begin{tabular}{lrrrr c lrrrr}
\toprule
\multicolumn{5}{c}{\textit{Positive}} & & \multicolumn{5}{c}{\textit{Negative}} \\
\cmidrule(lr){1-5}\cmidrule(lr){7-11}
\textbf{Emotion} & \textbf{Prec} & \textbf{Rec} & $\boldsymbol{F_1}$ & \textbf{Support} & &
\textbf{Emotion} & \textbf{Prec} & \textbf{Rec} & $\boldsymbol{F_1}$ & \textbf{Support} \\
\midrule
joy        & 74.1 & 19.9 & 31.4 & 201 & & anger         & 67.8 & 71.5 & 69.6 & 253 \\
gratitude  & 95.3 & 36.4 & 52.7 & 225 & & sadness       & 76.6 & 63.8 & 69.6 & 174 \\
relief     & 42.2 & 95.9 & 58.6 & 221 & & disgust       & 88.6 & 61.6 & 72.7 & 164 \\
pride      & 67.5 & 82.5 & 74.3 & 166 & & embarrassment & 71.2 & 95.2 & 81.5 & 210 \\
excitement & 81.3 & 67.4 & 73.7 & 187 & & fear          & 90.7 & 87.9 & 89.3 & 199 \\
\bottomrule
\end{tabular}
\caption{Per-emotion precision, recall, $F_1$, and support for GPT-5.5 on the full 1{,}000-sentence release of \textsc{Chiaro} against the adjudicated human gold. \emph{Relief} and \emph{embarrassment} are over-predicted, while \emph{joy} and \emph{gratitude} are under-predicted; the missed cases fall predominantly into \emph{relief}.}
\label{tab:llm-per-emotion}
\end{table*}

\paragraph{Annotation and adjudication.} Two fluent English-speaking annotators independently labeled all 1{,}050 generated sentences through a web-based interface, selecting one emotion per agent from the ten-class taxonomy (five positive--negative labels each). Annotators skipped items that failed quality standards, resulting in 1{,}017 scenes in the paired pool. We compute Cohen’s $\kappa$ separately for each polarity slot and obtain an average inter-annotator agreement of $\bar{\kappa} = 0.827$, with $\kappa_{\text{pos}} = 0.798$ (raw agreement $83.9\%$) and $\kappa_{\text{neg}} = 0.855$ (raw agreement $88.5\%$). We use these pre-adjudication agreements as the human reference point for model comparison. For items where the two annotators initially disagreed, they jointly discussed the appraisal cues and resolved disagreements within the same 1{,}017-sentence pool to produce a single gold label per agent. We randomly sample 1{,}000 as the final released dataset. We conduct all experiments on this set. The annotator instructions and interface are shown in Appendix~\ref{app:annotation-instructions}.

\paragraph{Dataset statistics.}\textsc{Chiaro} contains 1{,}000 sentences, each with adjudicated gold labels for both agents from two annotators. Although every scene is generated in two paired causal modes (a physical and a non-physical version), only one randomly chosen version per scene is annotated and released; the released corpus is therefore a per-scene random sample of the two. We release one version per scene for multiple reasons. The two versions of a scene share the same underlying story and characters, so releasing both would make roughly half the benchmark near-duplicates of the other half and allow information to leak between items. Secondly, labeling both versions would have halved scene coverage under our annotation budget (500 scenes with both versions vs.\ 1{,}000 scenes with one version). Moreover, a random pick per scene still preserves a fair comparison (526 physical vs.\ 474 non-physical).

Table~\ref{tab:stats} reports the per-emotion frequency within each polarity. The annotator-judgment distribution is approximately balanced within each polarity, with every class accounting for $16\%$--$25\%$.

\section{State-of-the-Art LLMs}
\label{sec:expLLMs}

We benchmark seven LLMs on \textsc{Chiaro}: OpenAI \texttt{gpt-5.5} \citep{openai2026gpt55}, Alibaba \texttt{Qwen3.6-Plus} \citep{qwen2026qwen36plus}, DeepSeek \texttt{V4-Pro} \citep{deepseek2026v4}, Meta \texttt{Llama-3.3-70B-Instruct} \citep{meta2024llama33}, Google \texttt{gemini-3.5-flash} \citep{deepmind2026gemini35flash}, Alibaba \texttt{Qwen3.5-27B} \citep{qwen2026qwen35_27b}, and Alibaba \texttt{Qwen3.5-9B} \citep{qwen2026qwen35_9b}. The first five are accessed through provider APIs; \texttt{Qwen3.5-27B} and \texttt{Qwen3.5-9B} are open-weights references at two scales. For every agent slot in every sentence, the model receives a polarity-filtered five-option choice set, the same form shown to human annotators in the \textsc{Chiaro} interface. All seven models use a \emph{joint} prompt that presents both agent role descriptions and both five-option choice sets in a single call and asks for two-letter answers. The full evaluation prompt is provided in Appendix~\ref{app:eval-prompt}.

Table~\ref{tab:llm-main} reports combined Agent-A and Agent-B macro-$F_1$ for the seven LLMs against the adjudicated human gold. GPT-5.5 leads with a macro-$F_1$ of $67.3$, followed by Qwen 3.6 Plus at $66.9$; the open-weights mid-scale Qwen3.5-27B sits at $66.3$ (tied with Llama 3.3 70B), while the smaller open Qwen3.5-9B trails at $59.9$. Inter-annotator agreement between the two annotators on the same paired pool is $\bar{\kappa}=0.827$ ($\kappa_{\text{pos}}=0.798$ on the positive slot, $\kappa_{\text{neg}}=0.855$ on the negative slot), which corresponds to 93.0 macro-F1 when the annotators are scored against the adjudicated gold. Therefore, even the strongest LLM sits roughly $26$ points below human agreement. We next dissect this gap along two axes: per-emotion errors (\S\ref{sec:per-emotion}) and the causal mode of the triggering event (\S\ref{sec:phys-nonphys}).

\begin{table}[t]
\centering\small
\setlength{\tabcolsep}{15pt}
\begin{tabular}{l S[table-format=2.1]}
\toprule
\textbf{Model} & \multicolumn{1}{r}{\textbf{Macro-$F_1$}} \\
\midrule
GPT-5.5                    & {\bfseries 67.3} \\
Qwen 3.6 Plus              & 66.9 \\
DeepSeek V4-Pro            & 66.5 \\
Qwen3.5-27B                & 66.3 \\
Llama 3.3 70B              & 66.3 \\
Gemini 3.5 Flash           & 64.1 \\
Qwen3.5-9B                 & 59.9 \\
\midrule
7-model mean               & 65.3 \\
\bottomrule
\end{tabular}
\caption{Macro-$F_1$ of the seven LLMs on the full 1{,}000-sentence release, scored against the adjudicated human gold. All evaluations use the joint two-agent prompt. Best value in bold. Per-emotion precision, recall, and $F_1$ breakdowns for the other six LLMs are in Appendix~\ref{app:per-emotion-all-llms}.}
\label{tab:llm-main}
\end{table}

\subsection{Per-emotion Error Analysis}
\label{sec:per-emotion}

Table~\ref{tab:llm-per-emotion} reports per-emotion precision, recall, and $F_1$ for GPT-5.5, the highest-performing LLM. Errors concentrate on the positive side, where two emotions are predicted more than the others. \emph{Relief} reaches $95.9\%$ recall at only $42.2\%$ precision, and \emph{embarrassment} reaches $95.2\%$ recall at $71.2\%$ precision. The corresponding deficits fall on \emph{joy} and \emph{gratitude}, whose recall drops to $19.9\%$ and $36.4\%$, respectively; most of the missed cases are mislabeled as \emph{relief}. 

Negative polarity shows a similar imbalance, but more weakly. \emph{Embarrassment} absorbs a portion of true \emph{anger}, \emph{sadness}, \emph{disgust}, and \emph{fear}. Per-class $F_1$ ranges from $31.4$ on \emph{joy} to $89.3$ on \emph{fear}, and the three lowest-$F_1$ emotions all sit on the positive side. The positive subset, therefore, accounts for most of the gap between GPT-5.5 and human agreement, suggesting that current LLMs struggle most with fine-grained positive emotions when explicit affect cues are removed. The two human annotators follow a similar ordering: their per-label agreement-$F_1$ is lowest on \emph{joy} ($78.3$) and highest on \emph{fear} ($94.0$). A plausible reason is the event’s specificity. \emph{Fear}’s mandatory trigger (an active, unresolved threat) is highly distinctive, whereas \emph{joy} acts as the default positive reading that competes with every other positive emotion.

\subsection{Physical vs Non-physical Causal Modes}
\label{sec:phys-nonphys}

Table~\ref{tab:llm-phys-nonphys} splits the macro-$F_1$ panel by the causal mode of the triggering event. Every API-served model scores 3--6 macro-$F_1$ points \emph{higher} on non-physical scenes than on physical ones (e.g., GPT-5.5 reaches $64.9$ points on physical vs $70.0$ on non-physical), contrary to the intuition that physical scenes should be easier because their trigger is concrete and observable. The mid-scale open Qwen-3.5-27B follows the same direction ($64.6$ vs $68.0$ points), while the smaller open Qwen-3.5-9B shows the opposite ordering ($60.8$ vs $58.6$ points). The pattern across larger models suggests that LLMs are as good as, if not better at, theory-of-mind-style inference about what each agent knows or perceives than at direct physical contact triggers. Human annotators show the same direction, where pre-adjudication agreement is higher on non-physical scenes ($\bar{\kappa}=0.849$) than on physical ones ($\bar{\kappa}=0.806$), suggesting the difficulty gap is intrinsic to the scenes rather than a model's output.

\begin{table}[t]
\centering\small
\setlength{\tabcolsep}{6pt}
\begin{tabular}{lrr}
\toprule
\textbf{Model} & \textbf{Physical} & \textbf{Non-physical} \\
\midrule
Qwen 3.6 Plus              & 65.4 & \textbf{68.5} \\
GPT-5.5                    & 64.9 & \textbf{70.0} \\
DeepSeek V4-Pro            & 64.4 & \textbf{68.5} \\
Qwen3.5-27B                & 64.6 & \textbf{68.0} \\
Llama 3.3 70B              & 64.5 & \textbf{68.3} \\
Gemini 3.5 Flash           & 61.6 & \textbf{66.6} \\
Qwen3.5-9B                 & \textbf{60.8} & 58.6 \\
\midrule
7-model mean               & 63.7 & \textbf{66.9} \\
\bottomrule
\end{tabular}
\caption{Macro-$F_1$ (\%) of the seven LLMs on \textsc{Chiaro} split by causal mode of the triggering event (526 physical scenes vs 474 non-physical scenes), measured against the adjudicated human gold.}
\label{tab:llm-phys-nonphys}
\end{table}

\section{Emotion Classifiers}
\label{sec:classifiers}
\begin{figure*}[t]
\centering
\includegraphics[width=0.95\textwidth]{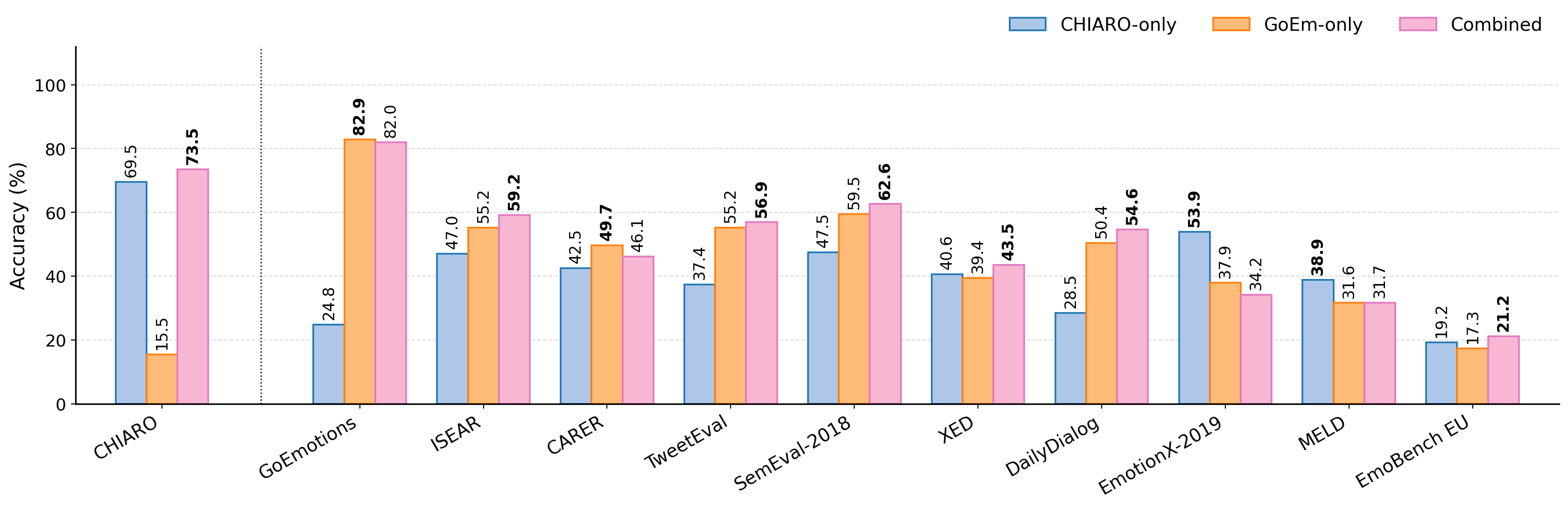}
\caption{Transfer accuracy (\%) of the three RoBERTa-large checkpoints (CHIARO-only, GoEm-only, Combined) on the \textsc{Chiaro} held-out test split and ten external emotion benchmarks. Best of three checkpoints are in bold.}
\label{fig:transfer-bars}
\end{figure*}

Beyond frontier LLMs, we evaluate smaller dedicated emotion classifiers in two regimes. We first test whether four off-the-shelf encoder checkpoints trained on single-agent emotion corpora transfer to \textsc{Chiaro}'s two-agent attribution setting. We then ask whether \textsc{Chiaro} works as a training signal on its own and when combined with an existing emotion corpus. To answer this, we fine-tune three RoBERTa-large checkpoints and evaluate each on \textsc{Chiaro} and ten external emotion benchmarks.

\subsection{Off-the-Shelf Emotion Classifiers}
\label{sec:expEncoders}

We evaluate four off-the-shelf encoder checkpoints on the full 1{,}000-sentence release. These are ModernBERT-large~\citep{modernbert} and ModernBERT-base fine-tuned on GoEmotions~\citep{JdFE2025b}, the Emo Pillars contextless RoBERTa-large checkpoint \citep{shvets-2025-emo}, and Emollama-chat-7B \citep{liu-etal-2024-emollms}, an emotion-tuned chat model trained on an affective analysis instruction dataset. The first three are classifiers over the GoEmotions 28-class label space, and are scored with polarity-restricted argmax over the five \textsc{Chiaro} emotions in the gold’s polarity bucket. No alias mappings are used, so the encoder must address each \textsc{Chiaro} emotion by its exact label. We use Emollama-chat-7B as a generative emotion classifier, queried with the same joint MCQ prompt as the LLMs (\S\ref{sec:expLLMs}) and scored on the letter it returns for each slot.

Table~\ref{tab:expA} reports macro-$F_1$ against the adjudicated human gold. All four encoders score well below the LLMs, and the three GoEmotions-trained encoders span $11.8$--$29.0$ macro-$F_1$, sitting $36$--$54$ points below the LLM mean despite covering each \textsc{Chiaro} emotion as an exact label. Existing single-individual emotion classifiers therefore transfer to \textsc{Chiaro}'s contrastive agent-attributed setting at chance levels.

\begin{table}[t]
\centering\small
\setlength{\tabcolsep}{4pt}
\begin{tabular}{l S[table-format=2.1]}
\toprule
\textbf{Model} & \multicolumn{1}{r}{\textbf{Macro-$F_1$}} \\
\midrule
Emo Pillars                    & 11.8 \\
ModernBERT-base (GoEmotions)   & 21.4 \\
ModernBERT-large (GoEmotions)  & 29.0 \\
Emollama-chat-7B               & 28.1$^{\ast}$ \\
\midrule
7-LLM mean                     & \bfseries 65.3 \\
\bottomrule
\end{tabular}
\caption{Macro-$F_1$ of off-the-shelf emotion classifiers on the full 1{,}000-sentence \textsc{Chiaro} release vs the adjudicated human gold. \textsuperscript{$\ast$}Emollama-chat-7B is scored on 1{,}977 of 2{,}000 slots; 23 unparseable outputs are excluded.}
\label{tab:expA}
\end{table}

To understand why the GoEmotions-trained encoders fall so far behind the LLMs, we compare their behavior on \textsc{Chiaro} against their behavior on the source task they were trained for. On the GoEmotions test split filtered to \textsc{Chiaro}’s ten emotions, ModernBERT-large reaches $79.6$ macro-$F_1$, and ModernBERT-base reaches $76.9$ under identical scoring. They drop to $29.0$ and $21.4$ macro-$F_1$ on \textsc{Chiaro}, a $51$--$56$ point gap with architecture, scoring, and label set held fixed. The difference is per-agent attribution, i.e., single-agent classifiers measure emotion by text expression, while \textsc{Chiaro} requires the emotion attributed to a referenced individual inside the text.

\subsection{\textsc{Chiaro} as a Training Signal}
\label{sec:expRoBERTa}

Our final study asks whether \textsc{Chiaro} is a usable training signal on its own and whether combining it with an existing emotion dataset, such as GoEmotions, yields a stronger classifier than either source alone. We train three RoBERTa-large checkpoints that differ only in their training corpus and evaluate all three on the \textsc{Chiaro} test split and ten external emotion benchmarks. \emph{CHIARO-only} is fine-tuned on \textsc{Chiaro} under an 80-10-10 train-val-test split, with each sentence converted into two pair-input examples by pairing it with each agent’s role (1{,}600 training examples) so the model conditions on the target agent. \emph{GoEm-only} is a same-architecture baseline trained on 1{,}600 GoEmotions items. We sample up to 160 items per \textsc{Chiaro} emotion. Rare classes such as \emph{pride} and \emph{relief} have fewer than 160 items, so we fill the remainder from the more frequent classes. GoEmotions has no agent slot, so the input is the utterance alone.

\emph{Combined} is trained on the union of both corpora ($3{,}200$ examples), keeping each source’s native input shape. For each external item, we pair the utterance with ``the speaker'' and keep only items whose gold is one of \textsc{Chiaro}'s ten emotions. The ten external benchmarks are GoEmotions \citep{demszky-etal-2020-goemotions}, ISEAR \citep{scherer-wallbott-1994-isear}, CARER \citep{saravia-etal-2018-carer}, TweetEval \citep{barbieri-etal-2020-tweeteval}, SemEval-2018 Affect-in-Tweets \citep{mohammad-etal-2018-semeval}, XED \citep{ohman-etal-2020-xed}, DailyDialog \citep{li-etal-2017-dailydialog}, EmotionX-2019 \citep{shmueli-ku-2019-socialnlp}, MELD \citep{poria-etal-2019-meld}, and EmoBench EU \citep{sabour-etal-2024-emobench}; training hyperparameters are listed in Appendix~\ref{app:training}.

Figure~\ref{fig:transfer-bars} reports top-1 accuracy. The CHIARO-only checkpoint achieves $69.5\%$ on the \textsc{Chiaro} held-out test split, whereas RoBERTa-base on the same training data achieves only $44.0\%$; the 25-percentage-point gap confirms that the task is learnable but capacity-dependent. \emph{Combined} uniquely beats both single-source baselines on six of the ten external datasets and beats CHIARO-only in-distribution as well ($73.5\%$ vs $69.5\%$).

The remaining four splits factor cleanly. On EmotionX-2019 and MELD, the CHIARO-only checkpoint stays ahead by a wide margin, and adding GoEmotions data lowers accuracy. Both benchmarks are dialogue-style and ask for the emotion attributed to a specific speaker, which is closer to \textsc{Chiaro}'s task than to GoEmotions'. On GoEmotions and CARER, the GoEm-only checkpoint has a slightly higher accuracy; both are short texts with first-person expressed emotion, close to GoEm-only's training distribution. On EmoBench EU, \emph{Combined} again leads the three, consistent with the overall pattern, though the small sample keeps that lead suggestive rather than decisive. Taken together, this gives a clear understanding of when combining the training dataset helps. \textsc{Chiaro} adds an attribution signal that GoEmotions lacks. Similarly, GoEmotions adds a first-person text-expressed-emotion signal that \textsc{Chiaro} lacks; and the union is the strongest source whenever the test benchmark mixes both demands. This pattern shows that \textsc{Chiaro} and GoEmotions are complementary training signals rather than redundant ones, where \textsc{Chiaro} captures third-person, individual-attributed emotion and GoEmotions captures first-person, text-expressed emotion.

\section{Conclusion}

We introduced \textsc{Chiaro}, a 1{,}000-sentence human-annotated benchmark for two-agent contrastive emotion inference. Each scene presents opposite-valence emotions tied to a single causal event, with inference grounded in appraisal theory rather than affect vocabulary. Seven frontier LLMs reach a 7-model mean of $65.3$ macro-$F_1$, well below the $93.0$ macro-$F_1$ that the human annotators reach ($\bar{\kappa}=0.827$ inter-annotator agreement), leaving a substantial gap concentrated on the positive subset and on physically triggered scenes. Four off-the-shelf emotion classifiers transfer to \textsc{Chiaro} at near-chance level despite covering each of its ten labels exactly, isolating per-agent attribution as the missing piece. A RoBERTa-large fine-tuned on \textsc{Chiaro} alone reaches $69.5\%$ in-distribution accuracy. The union of \textsc{Chiaro} with a matched-size GoEmotions slice beats either source alone on \textsc{Chiaro} and on six of ten external emotion benchmarks. Together these findings position contrastive agent-attributed emotion as a distinct task family that current emotion resources lack, and that \textsc{Chiaro} fills as a complementary training signal.

\section*{Limitations}

\textsc{Chiaro} is English-only and its narrative grounding is drawn from a single online community (\emph{r/AmItheAsshole}), so the situations, social norms, and interpersonal scripts it spans are skewed toward U.S.\ and Anglophone framings. Generation is performed by a single model (\texttt{gpt-5.2}); even with lexical constraints, repair loops, and human adjudication, the dataset is likely to inherit residual stylistic and topical biases from that generator.  The annotation pool could be strengthened with a larger and demographically more diverse pool, which would tighten human agreement and reduce annotation biases. Finally, the task is restricted to a positive--negative valence pair, so the benchmark does not measure model behavior on same-polarity-but-different-emotion cases (e.g., two agents both feeling distinct negative emotions), nor on cases where one or both agents are emotionally neutral. Relatedly, \textsc{Chiaro} is sized as an evaluation benchmark and a complementary training signal rather than a deployment-scale training corpus. The released generation pipeline supports scaling the corpus and extending it to source communities beyond AITA, which we view as the natural next step.

\section*{Acknowledgments}

We thank the CincyNLP group for their suggestions and feedback. We also thank the anonymous EMNLP reviewers for
their insightful suggestions.
\bibliography{custom}

\appendix

\section{AITA Keyword-Based Post Selection}
\label{app:aita-keywords}

For each generated instance, we select an AITA post that contains keyword cues compatible with the target emotion pair (e.g., posts containing ``thanked'' or ``helped me'' for gratitude-targeted generation; posts containing ``in front of'' or ``publicly'' for embarrassment-targeted generation). Keyword filtering improves hit-rate during generation but is not enforced downstream; the language model is free to abstract the post into a different setting. Each AITA post is used at most once across the dataset.

\section{Generation Prompts and Decoding Hyperparameters}
\label{app:prompts}

All generation is performed against OpenAI \texttt{gpt-5.2}. The Stage~1 draft call uses temperature $1.0$; the Stage~2 render call uses temperature $0.8$; JSON-schema enforcement is applied at both stages. Generation is parallelised with a 10-worker thread pool, and the repair loop (\S\ref{sec:validation}) is capped at four retries before a version is discarded.

Prompts~B.1 and~B.2 generate the backbone draft scene with two contrasting agents.

\begin{promptbox}{Prompt B.1: Stage 1 Draft --- System Message (\texttt{DRAFT\_PROMPT\_TEMPLATE})}
You are writing a short draft scene involving exactly two human agents.

\textbf{Constraints:}
\begin{itemize}[leftmargin=*,itemsep=0pt,topsep=2pt]
\item Avoid stealing or creating villains.
\item The emotions of the two agents are OPPOSITE in valence (one positive, one negative).
\item Keep the language simple.
\item The sentence must make clear WHY each agent feels the way they do.
\end{itemize}

\textbf{EMOTION TAXONOMY} — each emotion has a MANDATORY TRIGGER that MUST appear in the story (full per-emotion trigger list as in Appendix~\ref{app:triggers}).

\{emotion\_guidance\}\pnote{mandatory-trigger block for the target $(e_+, e_-)$ pair}

\textbf{CONTRASTIVE SCENARIO TYPE:} \{category\_block\}\pnote{one of the six types in Appendix~\ref{app:scenarios}}

Return JSON with: \texttt{setting}, \texttt{agent\_A\_role}, \texttt{agent\_B\_role}, \texttt{draft\_story}.
\end{promptbox}

\begin{promptbox}{Prompt B.2: Stage 1 Draft --- User Message}
Story inspiration:\textbackslash n\{aita\_post\}\pnote{the selected AITA post, truncated to its first 2{,}000 characters}
\end{promptbox}

The template above is the \emph{system} message of the Stage~1 call; the selected AITA post is passed separately as the \emph{user} message, prefixed with ``Story inspiration:''. The post therefore seeds the scene without being part of the fixed template, and the generator is free to abstract it into a different setting (\S\ref{sec:pipeline}).

\paragraph{Stage 2 render core (\texttt{\_RENDER\_CORE}).} Shared by both physical and non-physical renders. The key rules are:
\begin{itemize}[leftmargin=*,itemsep=2pt,topsep=2pt]
\item \textbf{Single-cause rule.} A single event must cause both agents' emotions. \texttt{cause\_span} must be a phrase in the sentence that names this shared event; \texttt{evidence\_A} and \texttt{evidence\_B} must each be consequences of that same event, never two different triggers.
\item \textbf{Neutral-cause rule.} The shared event must be a neutral external happening or third-party action whose fallout affects A and B differently. The cause must \emph{not} be Agent A deliberately acting against Agent B to hurt, punish, or extract from them — adversarial framings turn A's ``positive'' emotion punitive.
\item \textbf{Required emotion contrast.} Exactly one positive and one negative emotion per version.
\item \textbf{Disambiguation rules.} The mandatory trigger from Appendix~\ref{app:triggers} for each agent's emotion must be visible in the sentence (e.g., \emph{relief} must show a prior threat; \emph{embarrassment} must show a public audience).
\item \textbf{Self-check.} The generator is instructed to re-read its output and rewrite if (a) the two emotions do not share a trigger, (b) A's action directly targets B, or (c) a more specific emotion from the taxonomy fits better than the chosen one (e.g., \emph{joy} when the trigger lacks a prior threat).
\item \textbf{Agent-role rule.} Agent A and Agent B must have distinct role descriptions that reference the same identifier used in the sentence (name or distinguishing trait); no introducing a relationship label (``Fianc\'e,'' ``Roommate'') that does not appear in the sentence.
\item \textbf{Natural-language rules.} Present tense, self-contained, no emotion words or behavioural cues, no dialogue or text on screens.
\end{itemize}

\begin{promptbox}{Prompt B.3: Physical Render (\texttt{PHYS\_RENDER\_PROMPT})}
Convert a DRAFT scene into the PHYSICAL VERSION only.

VERSION: \texttt{physical\_version} — a concrete physical action or object change is the shared cause (e.g., last item grabbed from a shelf, door closed, pan pulled from oven, key handed over). The action need not be Agent A acting against Agent B — a neutral party, a mechanism, or even Agent B can be the one performing it. What matters is that the physical event produces different outcomes for both agents.

\textit{[followed by} \texttt{\_RENDER\_CORE}\textit{]}
\end{promptbox}

\begin{promptbox}{Prompt B.4: Non-Physical Render (\texttt{NONPHYS\_RENDER\_PROMPT})}
Convert a DRAFT scene into the NON-PHYSICAL VERSION only.

VERSION: \texttt{non\_physical\_version} — the cause is a situational or contextual cue, not direct physical impact (e.g., a closed sign, an announcement, an empty shelf).

\textit{[followed by} \texttt{\_RENDER\_CORE}\textit{]}
\end{promptbox}

\paragraph{Mandatory-emotions extension.} For the balanced-sampling driver (\texttt{generate\_stories\_balanced.py}), the per-class quota is enforced by prepending a forced-target block to the draft prompt that names the exact $(e_+, e_-)$ pair the next scene must instantiate. This is the mechanism that produces the balanced distribution in Appendix~\ref{app:per-round-stats}.

\paragraph{Repair prompt (\texttt{\_REPAIR\_SYSTEM}).} Triggered when any validator from Appendix~\ref{app:validators} fails. Instructs the generator to rewrite the sentence so that (i) the shared event is neutral and not A acting against B; (ii) a single event causes both emotions; (iii) exactly one positive and one negative emotion are present; (iv) Agent A is positive, Agent B negative; (v) no banned words or behavioural cues appear; (vi) Agent A and B have distinct role descriptions that uniquely identify each person and use the same identifier as the sentence; (vii) the literal strings ``Agent A'' and ``Agent B'' do not appear in the sentence text.

\section{Contrastive Scenario Types}
\label{app:scenarios}

Each generated scene is drafted under one of six contrastive scenario types, encoding different structural forms of the cause--effect relation linking the two agents. The closest existing umbrella taxonomy is the \emph{fortunes-of-others} branch of the OCC model \citep{ortony1988cognitive}, which classifies emotions about other agents' outcomes along a 2$\times$2 of (event desirable / undesirable for the other) $\times$ (rater's pleasure / displeasure). Our six types refine that grid by additionally specifying the \emph{causal structure} linking the two agents' outcomes; each type is anchored below in a distinct literature.

\paragraph{Zero-sum gain/loss \citep{festinger-1954-comparison,smith-kim-2007-envy}.}
A single scarce resource is split such that one agent gains it and the other is denied; the two outcomes are mutually exclusive by construction. The emotional contrast is grounded in social-comparison theory: an agent's gain becomes affectively charged for a comparable other who is denied the same good, producing envy or schadenfreude rather than parallel independent reactions.
\begin{itemize}
  \item One person gets the last ticket; the other arrives at an empty counter.
  \item One student sees an A on their paper; the other sees an F on theirs.
\end{itemize}

\paragraph{Side-effect spillover \citep{knobe-2003-intentional,coase-1960-social-cost}.}
One agent's positively-motivated activity is the source of a negative byproduct for the second agent. The first agent's emotion is justified on its own terms; the second agent's emotion arises from an unintended overflow. The structure mirrors the externality formalism in welfare economics, and recruits the lay-psychological asymmetry documented by the Knobe effect, namely that observers reliably treat negative side-effects as intentional even when foreseen but not pursued.
\begin{itemize}
  \item A child bounces in their airplane seat from excitement and keeps kicking the seat-back, bothering the passenger in front.
  \item A musician practices a new song in their apartment while the neighbor cannot concentrate.
\end{itemize}

\paragraph{Asymmetric information \citep{akerlof-1970-lemons,lazarus-1991-emotion}.}
The same event is experienced differently because the agents possess different knowledge or stakes. The contrast arises from the appraisal context, not from the physical event itself: each agent's goal-relevance and core relational theme differs, yielding divergent emotions over the same world-state.
\begin{itemize}
  \item A student learns they got early admission while their friend has not heard back.
  \item A worker finds out they passed probation while the colleague's contract will not be renewed.
\end{itemize}

\paragraph{Unintended consequence \citep{merton-1936-unanticipated,williams-1981-moral-luck}.}
The first agent acts with positive purpose, but an unintended downstream effect harms the second agent. Distinguished from \emph{side-effect spillover} by the requirement that the harm arises through a chain of events rather than as a direct byproduct of the action. The structure foregrounds the moral-luck asymmetry between A's appraisal (``I meant well'') and B's appraisal (``I was harmed''), where resultant luck drives the affective contrast.
\begin{itemize}
  \item A gardener waters flowers and the runoff floods the neighbor's mulch.
  \item A teacher rearranges seating for a reading corner, but one student loses their window seat.
\end{itemize}

\paragraph{Competing preferences \citep{deutsch-1973-conflict}.}
The two agents share an environment but have opposing needs along the same dimension; satisfying one preference automatically works against the other. The structure is the classical \emph{negative goal interdependence} of interdependence theory: pairs in which one agent's preferred environmental state precludes the other's.
\begin{itemize}
  \item A parent turns on the AC but their child was already cold.
  \item One roommate opens the window for a breeze while the other's papers blow off the desk.
\end{itemize}

\paragraph{Success vs.\ failure \citep{weiner-1985-attributional,wills-1981-downward}.}
Both agents independently attempt the same challenge; one succeeds while the other fails. The contrast is comparative rather than causally entangled (the two outcomes are produced by parallel, not interacting, paths), but the failure is affectively salient \emph{for the partner} via downward comparison, and attribution-theoretic appraisal of locus and controllability shapes the discrete emotions (pride, shame, pity) the two agents end up holding.
\begin{itemize}
  \item One runner finishes the marathon while another drops out from a cramp.
  \item One baker's souffl\'e rises perfectly while the other's collapses.
\end{itemize}

\paragraph{Type selection and final distribution.}
During the \textsc{Draft} stage, the scenario type is sampled uniformly at random from the six categories and inserted into the prompt as a structural constraint. The same scene draft is then rendered in both physical and non-physical versions, inheriting the sampled type. We do not enforce strict balance during sampling; small imbalances in the final corpus arise from variation in repair-loop retry counts across types. The realized distribution is: competing preferences $18.4\%$, zero-sum gain/loss $17.5\%$, side-effect spillover $17.0\%$, unintended consequence $16.9\%$, success vs.\ failure $15.1\%$, asymmetric information $15.0\%$.

\section{Lexical Constraint List}
\label{app:banned}

The validator at \S\ref{sec:validation} rejects any generated sentence containing a case-insensitive whole-word match against the list below ($n=71$). The list combines explicit emotion adjectives and their morphological variants, stereotyped facial-expression descriptors, and body-language phrases that frequently leak the target emotion:

\smallskip
\noindent\textbf{Affect adjectives and noun variants (forty-nine terms).}
\emph{happy, happily, joy, joyful, delighted; proud, pride, prideful; relieved, relief; grateful, gratitude, thankful; excited, excitement, eager, eagerly; sad, sadly, sorrow, sorrowful; angry, anger, furious, mad, enraged; fear, fearful, afraid, scared, terrified; guilty, guilt, regret, regretful; disgust, disgusted, disgusting, revolting, repulsed; embarrassed, embarrassment, ashamed, humiliated, shame; upset, annoyed, frustrated, dismayed.}

\smallskip
\noindent\textbf{Facial / body-language descriptors and bridging phrases (twenty-two terms).}
\emph{slumped shoulders, tight jaw, teary eyes, welling eyes; visibly, clearly, nervously; nods, smiles, frowns, glares, cries, screams, laughs, grins, claps; snatches, pumps, yanks; raises a fist, jumps for joy; throws up his hands.}

\noindent Per-emotion leakage extensions (e.g., \emph{cheers} for joy, \emph{sobs} for sadness) are applied on top of this base list during the validator's per-agent emotion-leakage check.

\section{Validation Checks}
\label{app:validators}

Every generated version (one per causal mode) is passed through six validators in sequence. A version that fails any check is returned to the generator with the explicit list of violations and asked to repair; the loop is bounded at four retries, after which the version is discarded. The validators are:

\begin{enumerate}[leftmargin=*,itemsep=2pt,topsep=2pt]
\item \textbf{Valence contrast.} Each scene must contain exactly one positive emotion (\emph{joy, pride, relief, gratitude, excitement}) and one negative emotion (\emph{anger, sadness, fear, disgust, embarrassment}), one per agent.

\item \textbf{Lexical constraints.} The sentence is rejected if it contains any case-insensitive whole-word match against the banned list of Appendix~\ref{app:banned}, or against the per-emotion leakage extension specific to either agent's gold emotion (e.g., \emph{cheers} leaks joy; \emph{sobs} leaks sadness).

\item \textbf{Length.} The sentence must be at most $300$ characters. Longer sentences are returned with a request to simplify while keeping both agents' outcomes clear.

\item \textbf{Person-reference count.} The sentence must have no more than two distinctly identified people driving the action (Agents A and B). Incidental mentions of other people can occur. Crowded scenes with additional named characters or referential ambiguity are rejected.

\item \textbf{Span consistency.} The \texttt{cause\_span}, \texttt{evidence\_A}, and \texttt{evidence\_B} fields produced by the generator must each be exact substrings of the final sentence, so that a single triggering event is grounded in the text.

\item \textbf{Role-head collision.} The two \texttt{agent\_role} strings must be distinguishable from their role prefixes alone (excluding common stopwords like ``the,'' ``who''). Identical roles or substantial token overlap is rejected, since the annotator must be able to tell A and B apart from the role description without re-reading the sentence.
\end{enumerate}

A separate single-cause heuristic, applied alongside the six validators, rejects scenes whose \texttt{cause\_effect\_relation} field describes two independent triggers (e.g., contains a ``while B's $\ldots$\ comes from $\ldots$'' construction, or two distinct \emph{because}-clauses) — a single event must produce both agents' outcomes.

\section{Training Hyperparameters}
\label{app:training}

For all three RoBERTa-large checkpoints in \S\ref{sec:expRoBERTa} (\emph{CHIARO-only}, \emph{GoEm-only}, \emph{Combined}), we use AdamW with learning rate $2 \times 10^{-5}$, weight decay $0.01$, batch size $16$, bf16 mixed precision, and $5$ epochs with best-on-validation-accuracy checkpoint selection. All three checkpoints are trained with seed $42$. RoBERTa-large has approximately $355$M parameters. Training was performed on a single NVIDIA A100 GPU; each checkpoint completed in approximately 2--3 hours wall-clock.

\section{Taxonomy: Discard Mapping and Mandatory Triggers}
\label{app:triggers}

Table~\ref{tab:discard-mapping} maps each of the eighteen non-retained GoEmotions categories (\emph{neutral} included) to either a retained \textsc{Chiaro} emotion or to an outright drop, with the criterion that drove the decision. Table~\ref{tab:triggers} gives the mandatory situational trigger paired with each of the ten retained emotions; the trigger is inserted into the Stage~1 draft prompt so the generated sentence has a recoverable disambiguation cue even under the lexical constraints of \S\ref{sec:validation}.

\begin{table}[ht]
\centering\small
\setlength{\tabcolsep}{4pt}
\begin{tabular}{lll}
\toprule
\textbf{GoEm category} & \textbf{Mapped to} & \textbf{Criterion} \\
\midrule
admiration       & gratitude   & appraisal overlap \\
amusement        & joy         & cluster dedup. \\
annoyance        & anger       & intensity sibling \\
approval         & gratitude   & appraisal overlap \\
caring           & gratitude   & appraisal overlap \\
desire           & excitement  & appraisal overlap \\
disappointment   & sadness     & appraisal overlap \\
disapproval      & disgust     & appraisal overlap \\
grief            & sadness     & intensity sibling \\
love             & gratitude   & appraisal overlap \\
nervousness      & fear        & intensity sibling \\
optimism         & excitement  & appraisal overlap \\
remorse          & embarrassment & appraisal overlap \\
\midrule
surprise         & \emph{dropped} & no fixed valence \\
curiosity        & \emph{dropped} & no fixed valence \\
realization      & \emph{dropped} & no fixed valence \\
confusion        & \emph{dropped} & no fixed valence \\
neutral          & \emph{dropped} & not an emotion \\
\bottomrule
\end{tabular}
\caption{Discard mapping from GoEmotions to \textsc{Chiaro}. ``Cluster dedup.''\ removes a near-synonym; ``intensity sibling'' removes a lower-/higher-intensity variant; ``appraisal overlap'' removes a category that occupies the same appraisal-theoretic cell as a retained one. Four GoEmotions categories plus \emph{neutral} are dropped outright for lacking a fixed valence.}
\label{tab:discard-mapping}
\end{table}

\begin{table}[ht]
\centering\small
\setlength{\tabcolsep}{4pt}
\begin{tabular}{lp{0.65\columnwidth}}
\toprule
\textbf{Emotion} & \textbf{Mandatory trigger} \\
\midrule
joy           & Agent receives or gains something good (the positive outcome has \emph{already} happened). \\
pride         & Agent accomplished something through their \emph{own} effort, skill, or work; personal achievement is shown. \\
relief        & A \emph{prior} threat or worry was avoided or resolved (without the threat, it is just joy). \\
gratitude     & Another person specifically helped, supported, or sacrificed for the agent; the helper is identifiable. \\
excitement    & Something good is \emph{about to} happen but has not yet; the agent looks forward to a future event. \\
anger         & Another person treated the agent unfairly, unjustly, or selfishly; a clear wrongdoer exists. \\
sadness       & Agent lost or was denied something, but \emph{no one is to blame} — it is circumstance or bad luck. \\
fear          & A bad outcome has \emph{not yet} happened but might; the threat is still active and unresolved. \\
disgust       & Someone did something morally revolting; the agent is repulsed by another person's behavior. \\
embarrassment & Agent was exposed, shamed, or failed with \emph{other people watching}; a public audience is present. \\
\bottomrule
\end{tabular}
\caption{Mandatory situational triggers paired with each of \textsc{Chiaro}'s ten emotions. Each trigger is inserted into the Stage~1 draft prompt and is enforced as a soft requirement during generation; the trigger is what lets a reader recover the intended emotion from a sentence that contains no affect vocabulary.}
\label{tab:triggers}
\end{table}

\section{Emotion Balance During Construction}
\label{app:per-round-stats}

\textsc{Chiaro} was generated in stages. Early generation used loose sampling: a target emotion pair was drawn uniformly, but the generator was not forced to respect the per-class budget, so the realized distribution was strongly skewed (\emph{disgust} $20.0\%$, \emph{gratitude} $17.0\%$; \emph{sadness} only $3.4\%$). We then added a per-class quota and a balanced-sampling driver that re-issues generation requests until each class hits its target, which brought every emotion within roughly $\pm 1.5$ percentage points of the uniform $10\%$ baseline. The released corpus, adjudicated down to $1{,}000$ scenes (Table~\ref{tab:stats}), inherits this balance with small drift from annotator skip and re-label decisions. Table~\ref{tab:per-round} shows the overall emotion distribution before balancing, after balancing, and in the final release.

\begin{table}[ht]
\centering\small
\setlength{\tabcolsep}{6pt}
\begin{tabular}{lrrr}
\toprule
\textbf{Emotion} & \textbf{Before} & \textbf{After} & \textbf{Release} \\
\midrule
joy           & 10.0 & 10.5 & 10.1 \\
pride         &  6.4 &  8.9 &  8.3 \\
relief        &  6.4 &  9.8 & 11.1 \\
gratitude     & 17.0 & 11.3 & 11.3 \\
excitement    & 10.2 &  9.5 &  9.4 \\
anger         & 10.2 & 10.8 & 12.7 \\
sadness       &  3.4 &  9.4 &  8.7 \\
fear          & 10.0 &  9.9 & 10.0 \\
disgust       & 20.0 & 10.2 &  8.2 \\
embarrassment &  6.4 &  9.6 & 10.5 \\
\bottomrule
\end{tabular}
\caption{Per-emotion frequency (\%, across both agent slots) before the balanced-sampling driver was introduced, after it, and in the final adjudicated release. The uniform baseline is $10\%$ per class.}
\label{tab:per-round}
\end{table}

\section{Annotator Instructions and Interface}
\label{app:annotation-instructions}

\begin{figure*}[t]
    \centering
    \includegraphics[width=0.95\textwidth]{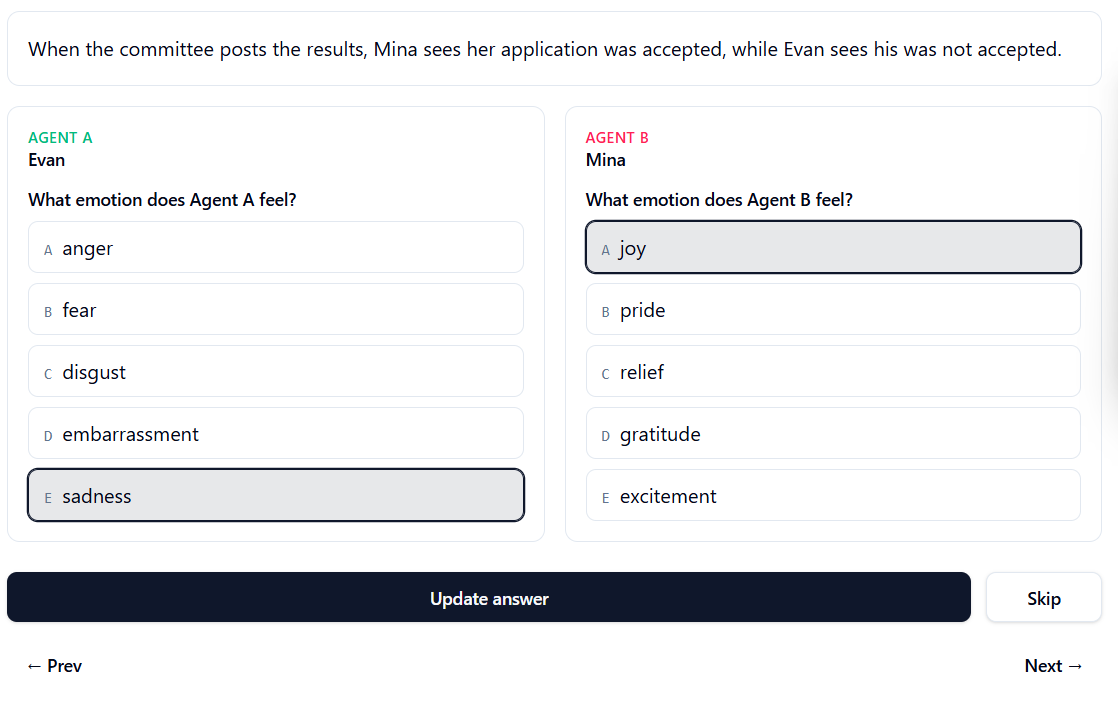}
    \caption{Screenshot of the \textsc{Chiaro} annotation interface.}
    \label{fig:annotation-interface}
\end{figure*}

Annotators were given the task description and labeling instructions through the web-based annotation interface described in \S\ref{sec:annotation}. Figure~\ref{fig:annotation-interface} shows a screenshot of the interface as presented to the two annotators.

\section{Per-Emotion Breakdown for All LLMs}
\label{app:per-emotion-all-llms}

Table~\ref{tab:llm-per-emotion} in \S\ref{sec:per-emotion} reports per-emotion precision, recall, and $F_1$ for GPT-5.5, the highest-performing model. We extend that breakdown to the remaining six LLMs in Tables~\ref{tab:per-emo-qwen36plus}--\ref{tab:per-emo-qwen35_9b}, all scored against the adjudicated human gold with predictions pooled across both agent slots. Across models, the positive subset (especially \emph{joy} and \emph{gratitude}) shows the widest variance and the lowest absolute $F_1$, while the negative subset is comparatively flat.

\begin{table*}[t]
\centering\small
\setlength{\tabcolsep}{8pt}
\begin{tabular}{lrrrr c lrrrr}
\toprule
\multicolumn{5}{c}{\textit{Positive}} & & \multicolumn{5}{c}{\textit{Negative}} \\
\cmidrule(lr){1-5}\cmidrule(lr){7-11}
\textbf{Emotion} & \textbf{Prec} & \textbf{Rec} & $\boldsymbol{F_1}$ & \textbf{Support} & &
\textbf{Emotion} & \textbf{Prec} & \textbf{Rec} & $\boldsymbol{F_1}$ & \textbf{Support} \\
\midrule
joy           & 60.3 & 40.8 & 48.7 & 201 & & anger         & 67.6 & 67.6 & 67.6 & 253 \\
gratitude     & 92.6 & 28.0 & 43.0 & 225 & & sadness       & 70.1 & 70.1 & 70.1 & 174 \\
relief        & 44.9 & 89.1 & 59.7 & 221 & & disgust       & 86.9 & 64.6 & 74.1 & 164 \\
pride         & 66.1 & 75.3 & 70.4 & 166 & & embarrassment & 70.5 & 92.4 & 80.0 & 210 \\
excitement    & 73.2 & 65.8 & 69.3 & 187 & & fear          & 92.0 & 81.4 & 86.4 & 199 \\
\bottomrule
\end{tabular}
\caption{Per-emotion precision, recall, $F_1$, and support for Qwen 3.6 Plus on the full 1{,}000-sentence release of \textsc{Chiaro} against the adjudicated human gold.}
\label{tab:per-emo-qwen36plus}
\end{table*}

\begin{table*}[t]
\centering\small
\setlength{\tabcolsep}{8pt}
\begin{tabular}{lrrrr c lrrrr}
\toprule
\multicolumn{5}{c}{\textit{Positive}} & & \multicolumn{5}{c}{\textit{Negative}} \\
\cmidrule(lr){1-5}\cmidrule(lr){7-11}
\textbf{Emotion} & \textbf{Prec} & \textbf{Rec} & $\boldsymbol{F_1}$ & \textbf{Support} & &
\textbf{Emotion} & \textbf{Prec} & \textbf{Rec} & $\boldsymbol{F_1}$ & \textbf{Support} \\
\midrule
joy           & 65.2 & 28.9 & 40.0 & 201 & & anger         & 68.0 & 69.6 & 68.8 & 253 \\
gratitude     & 92.5 & 32.9 & 48.5 & 225 & & sadness       & 67.9 & 63.2 & 65.5 & 174 \\
relief        & 42.6 & 95.0 & 58.8 & 221 & & disgust       & 86.7 & 63.8 & 73.5 & 164 \\
pride         & 65.3 & 78.3 & 71.2 & 166 & & embarrassment & 69.2 & 94.3 & 79.8 & 210 \\
excitement    & 84.1 & 62.4 & 71.6 & 187 & & fear          & 94.2 & 81.4 & 87.3 & 199 \\
\bottomrule
\end{tabular}
\caption{Per-emotion precision, recall, $F_1$, and support for DeepSeek-V4-Pro on the full 1{,}000-sentence release of \textsc{Chiaro} against the adjudicated human gold.}
\label{tab:per-emo-deepseekv4}
\end{table*}

\begin{table*}[t]
\centering\small
\setlength{\tabcolsep}{8pt}
\begin{tabular}{lrrrr c lrrrr}
\toprule
\multicolumn{5}{c}{\textit{Positive}} & & \multicolumn{5}{c}{\textit{Negative}} \\
\cmidrule(lr){1-5}\cmidrule(lr){7-11}
\textbf{Emotion} & \textbf{Prec} & \textbf{Rec} & $\boldsymbol{F_1}$ & \textbf{Support} & &
\textbf{Emotion} & \textbf{Prec} & \textbf{Rec} & $\boldsymbol{F_1}$ & \textbf{Support} \\
\midrule
joy           & 68.5 & 31.3 & 43.0 & 201 & & anger         & 72.3 & 66.0 & 69.0 & 253 \\
gratitude     & 87.2 & 36.4 & 51.4 & 225 & & sadness       & 67.6 & 67.2 & 67.4 & 174 \\
relief        & 45.3 & 93.7 & 61.1 & 221 & & disgust       & 85.8 & 66.5 & 74.9 & 164 \\
pride         & 63.2 & 69.3 & 66.1 & 166 & & embarrassment & 65.8 & 93.3 & 77.2 & 210 \\
excitement    & 68.0 & 63.6 & 65.7 & 187 & & fear          & 94.2 & 80.9 & 87.0 & 199 \\
\bottomrule
\end{tabular}
\caption{Per-emotion precision, recall, $F_1$, and support for Qwen3.5-27B on the full 1{,}000-sentence release of \textsc{Chiaro} against the adjudicated human gold.}
\label{tab:per-emo-qwen3527b}
\end{table*}

\begin{table*}[t]
\centering\small
\setlength{\tabcolsep}{8pt}
\begin{tabular}{lrrrr c lrrrr}
\toprule
\multicolumn{5}{c}{\textit{Positive}} & & \multicolumn{5}{c}{\textit{Negative}} \\
\cmidrule(lr){1-5}\cmidrule(lr){7-11}
\textbf{Emotion} & \textbf{Prec} & \textbf{Rec} & $\boldsymbol{F_1}$ & \textbf{Support} & &
\textbf{Emotion} & \textbf{Prec} & \textbf{Rec} & $\boldsymbol{F_1}$ & \textbf{Support} \\
\midrule
joy           & 62.3 & 37.8 & 47.1 & 201 & & anger         & 72.9 & 62.8 & 67.5 & 253 \\
gratitude     & 77.8 & 40.4 & 53.2 & 225 & & sadness       & 57.6 & 74.1 & 64.8 & 174 \\
relief        & 49.6 & 88.2 & 63.5 & 221 & & disgust       & 86.3 & 68.9 & 76.6 & 164 \\
pride         & 65.9 & 72.3 & 69.0 & 166 & & embarrassment & 67.8 & 89.0 & 77.0 & 210 \\
excitement    & 62.4 & 62.0 & 62.2 & 187 & & fear          & 95.4 & 72.4 & 82.3 & 199 \\
\bottomrule
\end{tabular}
\caption{Per-emotion precision, recall, $F_1$, and support for Llama 3.3 70B on the full 1{,}000-sentence release of \textsc{Chiaro} against the adjudicated human gold.}
\label{tab:per-emo-llama33_70b}
\end{table*}

\begin{table*}[t]
\centering\small
\setlength{\tabcolsep}{8pt}
\begin{tabular}{lrrrr c lrrrr}
\toprule
\multicolumn{5}{c}{\textit{Positive}} & & \multicolumn{5}{c}{\textit{Negative}} \\
\cmidrule(lr){1-5}\cmidrule(lr){7-11}
\textbf{Emotion} & \textbf{Prec} & \textbf{Rec} & $\boldsymbol{F_1}$ & \textbf{Support} & &
\textbf{Emotion} & \textbf{Prec} & \textbf{Rec} & $\boldsymbol{F_1}$ & \textbf{Support} \\
\midrule
joy           & 61.1 & 28.9 & 39.2 & 201 & & anger         & 66.4 & 71.1 & 68.7 & 253 \\
gratitude     & 92.5 & 27.6 & 42.5 & 225 & & sadness       & 67.3 & 66.1 & 66.7 & 174 \\
relief        & 37.7 & 95.5 & 54.1 & 221 & & disgust       & 87.2 & 57.9 & 69.6 & 164 \\
pride         & 67.3 & 60.8 & 63.9 & 166 & & embarrassment & 73.3 & 95.2 & 82.8 & 210 \\
excitement    & 80.6 & 55.6 & 65.8 & 187 & & fear          & 93.8 & 82.9 & 88.0 & 199 \\
\bottomrule
\end{tabular}
\caption{Per-emotion precision, recall, $F_1$, and support for Gemini 3.5 Flash on the full 1{,}000-sentence release of \textsc{Chiaro} against the adjudicated human gold.}
\label{tab:per-emo-gemini35f}
\end{table*}

\begin{table*}[t]
\centering\small
\setlength{\tabcolsep}{8pt}
\begin{tabular}{lrrrr c lrrrr}
\toprule
\multicolumn{5}{c}{\textit{Positive}} & & \multicolumn{5}{c}{\textit{Negative}} \\
\cmidrule(lr){1-5}\cmidrule(lr){7-11}
\textbf{Emotion} & \textbf{Prec} & \textbf{Rec} & $\boldsymbol{F_1}$ & \textbf{Support} & &
\textbf{Emotion} & \textbf{Prec} & \textbf{Rec} & $\boldsymbol{F_1}$ & \textbf{Support} \\
\midrule
joy           & 47.1 & 27.9 & 35.0 & 201 & & anger         & 68.0 & 53.8 & 60.0 & 253 \\
gratitude     & 63.2 & 42.7 & 50.9 & 225 & & sadness       & 59.7 & 52.9 & 56.1 & 174 \\
relief        & 49.3 & 79.2 & 60.8 & 221 & & disgust       & 71.6 & 64.6 & 67.9 & 164 \\
pride         & 49.8 & 75.3 & 60.0 & 166 & & embarrassment & 57.7 & 89.0 & 70.0 & 210 \\
excitement    & 76.4 & 50.3 & 60.6 & 187 & & fear          & 82.8 & 72.4 & 77.2 & 199 \\
\bottomrule
\end{tabular}
\caption{Per-emotion precision, recall, $F_1$, and support for Qwen3.5-9B on the full 1{,}000-sentence release of \textsc{Chiaro} against the adjudicated human gold.}
\label{tab:per-emo-qwen35_9b}
\end{table*}

\section{Evaluation Prompt}
\label{app:eval-prompt}

All seven LLMs (and Emollama-chat-7B, \S\ref{sec:expEncoders}) receive the same multiple-choice prompt, shown below. The five options per agent are the five emotions of that agent's gold polarity, and the correct option letters are rotated across scenes.

\begin{promptbox}{Prompt K.1: System Message}
You are answering a multiple-choice question about a sentence that describes two people reacting to the same event with contrasting emotions.

Read the sentence carefully and select the single best answer for each agent.

Reply with EXACTLY two lines in this format:\\
AGENT A: <letter>\\
AGENT B: <letter>\\
Nothing else.
\end{promptbox}

\end{document}